\documentclass[letterpaper, 10 pt, conference]{ieeeconf}  

\IEEEoverridecommandlockouts                              

\usepackage[nolist]{acronym} 
\usepackage{amsfonts,siunitx,booktabs,cite} 
\usepackage{graphics} 
\usepackage{epsfig} 
\usepackage{mathptmx} 
\usepackage{times} 
\usepackage{amsmath} 
\usepackage{amssymb}  
\usepackage{stmaryrd}

\usepackage{graphicx}
\usepackage{caption}
\let\labelindent\relax
\usepackage{enumitem}
\usepackage{svg}
\usepackage{float}
\usepackage[normalem]{ulem}
\usepackage{hyperref}
\usepackage{algorithm}
\usepackage{algpseudocode}

\usepackage{scalerel}
\usepackage{tikz,standalone,pgfplots}

\usepackage{soul}
\newacro{dnn}[DNN]{Deep Neural Network}
\newacro{fcn}[FCN]{Fully Convolutional Network}
\newacro{sdf}[SDF]{Signed Distance Function}
\newacro{cnn}[CNN]{Convolutional Neural Network}
\newacro{gnn}[GNN]{Graph Neural Network}
\newacro{dl}[DL]{Deep Learning}
\newacro{ml}[ML]{Machine Learning}
\newacro{mc}[MC]{Monte Carlo}
\newacro{mlp}[MLP]{Multi-Layer Perceptron}
\newacro{dof}[DoF]{Degrees of Freedom}
\newacro{vae}[VAE]{Variational Autoencoder}
\newacro{cvae}[CVAE]{Conditional Variational Autoencoder}
\newacro{methodname}[VCGS]{Variational Constrained Grasp Sampler}
\newacro{fps}[FPS]{Farthest Point Sampling}
\newacro{tai}[TaI]{Target as Input}
\newacro{pca}[PCA]{Principal Component Analysis}
\newacro{pc}[PC]{Principal Component}
\newacro{auc}[AUC]{Area Under the Curve}

\newcommand{\equationref}[1]{\hyperref[#1]{Eq.~\ref*{#1}}}
\newcommand{\figref}[1]{\hyperref[#1]{Fig.~\ref*{#1}}}
\newcommand{\tabref}[1]{\hyperref[#1]{Table~\ref*{#1}}}
\newcommand{\secref}[1]{\hyperref[#1]{Section~\ref*{#1}}}
\newcommand{\algoref}[1]{\hyperref[#1]{Algorithm~\ref*{#1}}}

\newcommand{\norm}[1]{\left\lVert#1\right\rVert}

\newcommand{\matr}[1]{\mathbf{#1}}

\newcommand*{\prob}{\mathsf{P}}

\newcommand{\etal}[1]{#1 et al.}

\def\methodname{GoNet}

\def\datasetacronomy{Acroym}

\def\graspnet{GraspNet}

\def\bestcolor{(best viewed in color)}
\def\sota{state-of-the-art}

\def\pc{point cloud}
\def\pcs{point clouds}

\def\pointnet{PointNet++~\cite{qiPointNetDeepHierarchical2017a}}

\title{\LARGE \bf
\methodname{}: An Approach-Constrained Generative Grasp Sampling Network
}

\author{Zehang Weng, Haofei Lu, Jens Lundell, and Danica Kragic
\thanks{All authors are with the division of Robotics, Perception, and Learning (RPL) at KTH, Stockholm, Sweden. 
        {\tt\small \{zehang,haofeil,jelundel,dani\}@kth.se}}%
}

\makeatletter
\let\@oldmaketitle\@maketitle
\renewcommand{\@maketitle}{\@oldmaketitle
  \setcounter{figure}{0}
  \vspace{1em}
  \centering
      \includegraphics[width=\linewidth]{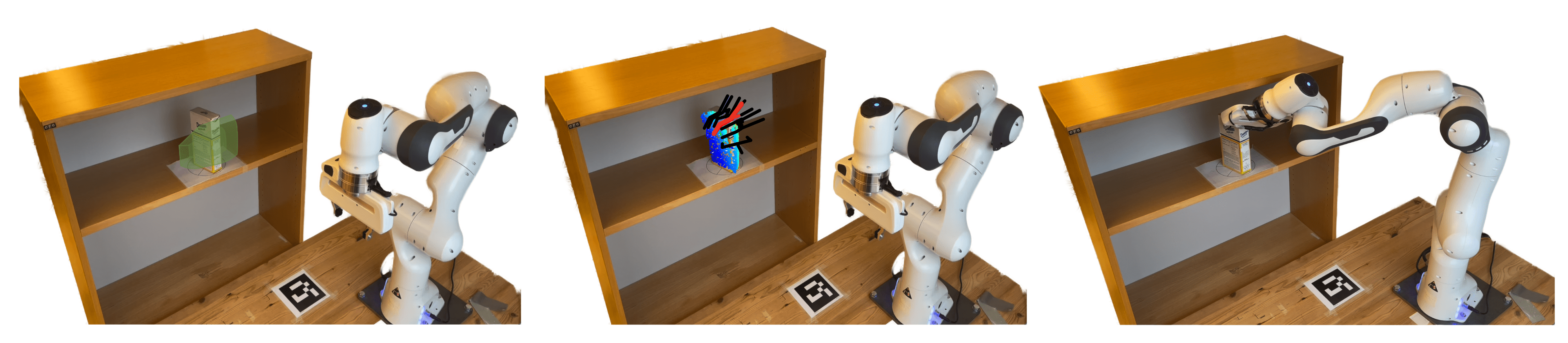}
      \captionof{figure}{The shelf-picking setup. Left: The green areas represent collision-free grasp directions. Middle: the grasps generated by our \methodname{} with the best grasp highlighted in red. Right: the robot successfully reaches the red grasp pose. \label{fig:shelf_exp_with_grasps}}
  }
\makeatother

\begin{document}
\maketitle
\thispagestyle{empty}
\pagestyle{empty}
\begin{abstract}

This work addresses the problem of learning approach-constrained data-driven grasp samplers. To this end, we propose \methodname{}: a generative grasp sampler that can constrain the grasp approach direction to a subset of $\mathrm{SO}(3)$. The key insight is to discretize $\mathrm{SO}(3)$ into a predefined number of bins and train \methodname{} to generate grasps whose approach directions are within those bins.  At run-time, the bin aligning with the second largest principal component of the observed \pc{} is selected. \methodname{} is benchmarked against \graspnet{}, a \sota{} unconstrained grasp sampler, in an unconfined grasping experiment in simulation and on an unconfined and confined grasping experiment in the real world. The results demonstrate that \methodname{} achieves higher success-over-coverage in simulation and a 12\%--18\% higher success rate in real-world table-picking and shelf-picking tasks than the baseline.

\end{abstract}

\section{introduction}

To date, many \sota{} data-driven grasp sampling methods have achieved high grasp success rates on completely unknown objects by learning distributions of successful grasps all over the object \cite{morrisonClosingLoopRobotic2018b,mahlerDexNetDeepLearning2017a,zhou2018fully,satish2019policy,kumra2020antipodal,zhu2022sample}. Although impressive, it is worth highlighting that the grasping conditions for these methods are very favorable as the workspace for the robot is often only restricted by the surface the robot and the object rest on. If the focus is instead shifted towards grasping from confined spaces, such as the example of picking from the shelf shown in \figref{fig:shelf_exp_with_grasps}, the grasp success rate falls. One reason the success rate falls when grasping an object resting on a shelf is that the often highly successful top-down approach direction is now impossible to reach, and the robot should instead grasp the object from the side.


In this work, we address the problem of generating approach-constrained grasps. Towards such a goal, we propose \methodname{}: the first \ac{dl}-based data-driven grasp sampler capable of generating approach-constrained grasps in all of $\mathrm{SO}(3)$, and a geometrical method to extract the specific grasp approach directions automatically from the object's \pc{}. To train \methodname{}, we first discretize $\mathrm{SO}(3)$ into a preset number of bins, each associated with a pitch-and-yaw angle. Then, from a given object \pc{} and bin, \methodname{} is trained to reconstruct successful grasps on that object with approach directions within that bin. Together, \methodname{} and the \ac{pc}-based bin selection approach can be seen as a geometrical approach-constrained data-driven grasp sampler. 

We empirically evaluate the performance of \methodname{} to the unconstrained \sota{} 6-\ac{dof} \graspnet{} \cite{mousavian20196} on over 17 million grasps in simulation, and over 100 grasps in two real-world experiments. The simulation results indicate that our method with the \ac{pc} inspired sampling direction generally achieves higher success-over-coverage than the baseline. Furthermore, in the real world, our method achieves a 12\% higher grasp success rate on a shelf-picking task and an 18\% higher grasp success rate on a table-picking task.   

To summarize, the main contributions of this work are:

\begin{itemize}
    \item \methodname{}: a novel 6-\ac{dof} generative grasp sampler that can constrain grasps to approach the object from specific directions in $\mathrm{SO}(3)$ (\secref{sec:generator}).
    \item A geometrically inspired approach for choosing the specific approach directions (\secref{sec:pc_selection}). 
    \item A dataset for training \methodname{} and a thorough discussion on the choice of coordinate systems for representing the grasps in the dataset (\secref{sec:dataset}). 
    \item Extensive simulation and real-world experiments of \methodname{} and \sota{} 6-\ac{dof} \graspnet{}, highlighting specific benefits and drawbacks of our constrained grasp sampler to an unconstrained sampler (\secref{sec:exp}). 
\end{itemize}

\section{related work}\label{sec:relatedwork}

Most constrained grasp sampling works either fall into task-based grasping, 4-\ac{dof} data-driven grasping, or geometrically constrained grasping. As these three categories are generally unrelated, we will review them separately.

\subsection{4-\ac{dof} Grasping}

4-\ac{dof} grasp samplers restrict the approach direction of the grasps to be perpendicular to a specific grasping plane \cite{morrisonClosingLoopRobotic2018b,mahlerDexNetDeepLearning2017a,zhou2018fully,satish2019policy,kumra2020antipodal,zhu2022sample}. As such, these methods are constrained by construction. One of the apparent benefits of constraining grasps to 4-\ac{dof} instead of 6 is that it restricts the space of possible grasps, which facilitates learning. If, additionally, grasps are constrained to a top-down direction, the surface the objects rest on hinders the gripper from tilting the object while approaching it, which generally improves grasp success rates substantially \cite{lundell2020beyond}. However, if grasps are generated from a camera plane that is not top-down, which is the case in this work, grasp success deteriorates drastically \cite{lundell2020beyond}. Therefore, we do not restrict the approach direction by construction and instead propose constraining it to \textit{any} subset of $\mathrm{SO}(3)$.


\subsection{Task-based Grasping}

As the name suggests, task-based grasp samplers address how to sample grasps for completing tasks. Early such approaches used numerical methods to find grasps that could achieve specific task wrenches \cite{borst2004grasp,haschke2005task}. Although these solutions are computationally efficient and mathematically sound, specifying a task in the abstract wrench space is non-trivial. To avoid the need for specifying task wrenches, later work focused on learning to sample task-specific grasps from data  \cite{murali2021same,antonovaGlobalSearchBernoulli2018a,kokic2017affordance,song2010learning,song2015task,fang2020learning,detry2017task}. The main difference between these data-driven methods is in the model for learning the task-specific grasps, which ranged from Bayesian networks \cite{song2010learning,song2015task}, kernel methods \cite{antonovaGlobalSearchBernoulli2018a}, to \acp{dnn} \cite{kokic2017affordance,murali2021same}. Compared to task-based grasping methods, our method is more flexible because it can generate grasps not only for completing tasks but also grasps that can, for example, avoid parts of the environment that the robot would otherwise risk colliding with.

\subsection{Geometrically Constrained Grasping}\label{sec:geometrically_constrained_grasping}

In geometrically constrained grasping, grasps are constrained to align with the geometry of the object model \cite{pas2016localizing,stoyanov2016grasp}. One of the primary reasons for aligning grasps to the geometry of the object was found in the seminal work by \etal{Balasubramanian} \cite{balasubramanian2012physical} where they experimentally demonstrated that in highly successful human-planned robotic grasps, the robot's wrist was mainly aligned to the first \acp{pc} of the object or the plane normal to the first \ac{pc}.

Our work does not explicitly constrain the sampled grasps to the object's geometry. Nevertheless, we do draw inspiration from these as we select bins that constrain the generated grasp directions to lie in the plane normal to the largest \ac{pc} and along the second largest \ac{pc} of the observed \pc{}.


\section{Problem Formulation}\label{sec:problem}

In this work, we address the problem of synthesizing successful (S=1) parallel-jaw grasps $\matr{G}$ on partially observed object point clouds $\matr{O} \in \mathbb{R}^{\text{N}\times 3}$ where the approach directions $\Vec{a} \in \mathbb{R}^3$ of the grasps are constrained to a subset $\text{C}\subset \mathrm{SO}(3)$ of the rotation group $\mathrm{SO}(3)$. Mathematically, the objective is to learn the joint distribution $\prob{(\matr{G}, \text{S}=1 | \matr{O}, \text{C})}$. 

We define a grasp as successful if it can pick and hold the target object. Based on that definition, $\matr{G}$ and $\matr{O}$ fully determine the grasp success probability. Thus, the success of a grasp $\text{S}$ is conditionally independent of $\text{C}$ given $\matr{G}$. Using this independence, we can factorize $\prob{(\matr{G}, \text{S}=1 | \matr{O}, \text{C})}$ into $\prob{(\text{S}=1 | \matr{G}, \matr{O})}\prob{(\matr{G}|\matr{O}, \text{C})}$, where the first distribution $\prob{(\text{S}=1 | \matr{G}, \matr{O})}$ is a grasp discriminator and the second $\prob{(\matr{G}|\matr{O}, \text{C})}$ is a grasp generator. We approximate these distributions with separate \acp{dnn} $\mathcal{Q}_{\boldsymbol{\theta}}(\matr{G}|\matr{O}, \text{C})\approx \prob{(\matr{G}|\matr{O}, \text{C})}$ and $\mathcal{D}_{\boldsymbol{\psi}}(\text{S}=1 | \matr{G}, \matr{O}) \approx \prob{(\text{S}=1 | \matr{G}, \matr{O})}$, each with their own trainable parameters $\boldsymbol{\theta}$ and $\boldsymbol{\psi}$. 

\begin{figure}[t]
\centering
  \centering
  \captionsetup{margin=0.25cm}
        \includegraphics[width=0.55\linewidth]{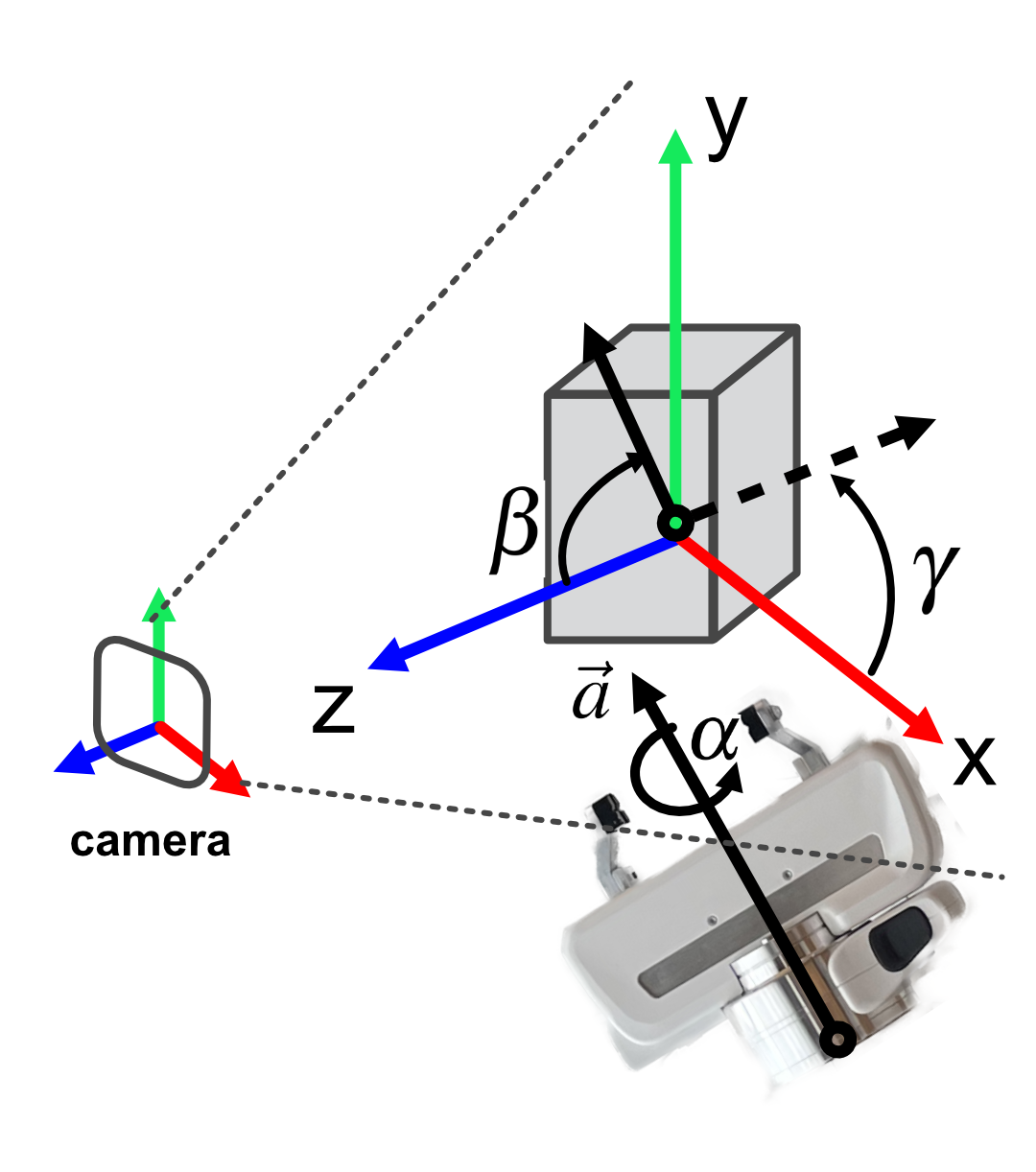}
    \captionof{figure}{The camera-centric grasp orientation representation. Here, $\Vec{a}$ is the gasp approach direction, $\alpha$ the roll angle, $\beta$ the pitch angle, and $\gamma$ the yaw angle, all represented in the camera coordinate system. 
    }
    \vspace{-9pt}
  \label{fig:yaw-pitch-cam}
\end{figure}

In this work, we represent grasp poses $\matr{G}=[\matr{R}, \matr{T}]\in \mathbb{R}^7$ by a rotation $\matr{R}$ expressed as a 4-D quaternion and a translation $\matr{T}$ represented as a 3-D vector. It is also possible to express the rotation $\matr{R}$ as a unit length approach vector $\Vec{a}\in \mathbb{R}^3$ and a hand orientation angle $\alpha$ or, in terms of Euler angles, as a roll ($\alpha$), pitch ($\beta$) and yaw ($\gamma$) angle, both of which are visualized in \figref{fig:yaw-pitch-cam}. Out of these components, the approach vector (pitch and yaw angles) is explicitly constrained by C, while the hand orientation (roll angle) is learned.

\section{Method}\label{sec:method}

As formalized in the previous section, two \acp{dnn} are required: a generative grasp sampler and a grasp discriminator. Next, we describe, in detail, \methodname{} our novel generative approach-constrained grasp sampling network and the automatic \ac{pc}-based bin selection process. Finally, we introduce the grasp discriminator from \cite{mousavian20196} for completeness but refer the reader to the original work in \cite{mousavian20196} for specific details.


\subsection{\methodname{}}\label{sec:generator}

\methodname{} is an approach-constrained deep generative grasp sampling network $\mathcal{Q}_{\boldsymbol{\theta}}(\matr{G}|\matr{O}, \text{C})$ that approximates $\prob{(\matr{G}|\matr{O}, \text{C})}$, where $\matr{O} \in \mathbb{R}^{\text{N}\times3}$ is the object \pc{} and $\text{C}\subset \mathrm{SO}(3)$ is a subset of the rotation group $\mathrm{SO}(3)$. As discussed in the previous section, the idea behind conditioning the network on $\text{C}$ is to constrain the approach vector $\Vec{a}$ or the pitch and yaw Euler angles to lie within some subset of $\mathrm{SO}(3)$.  

\begin{figure}[t]
  \centering
  \captionsetup{margin=0.25cm}
        \includegraphics[width=1\linewidth]{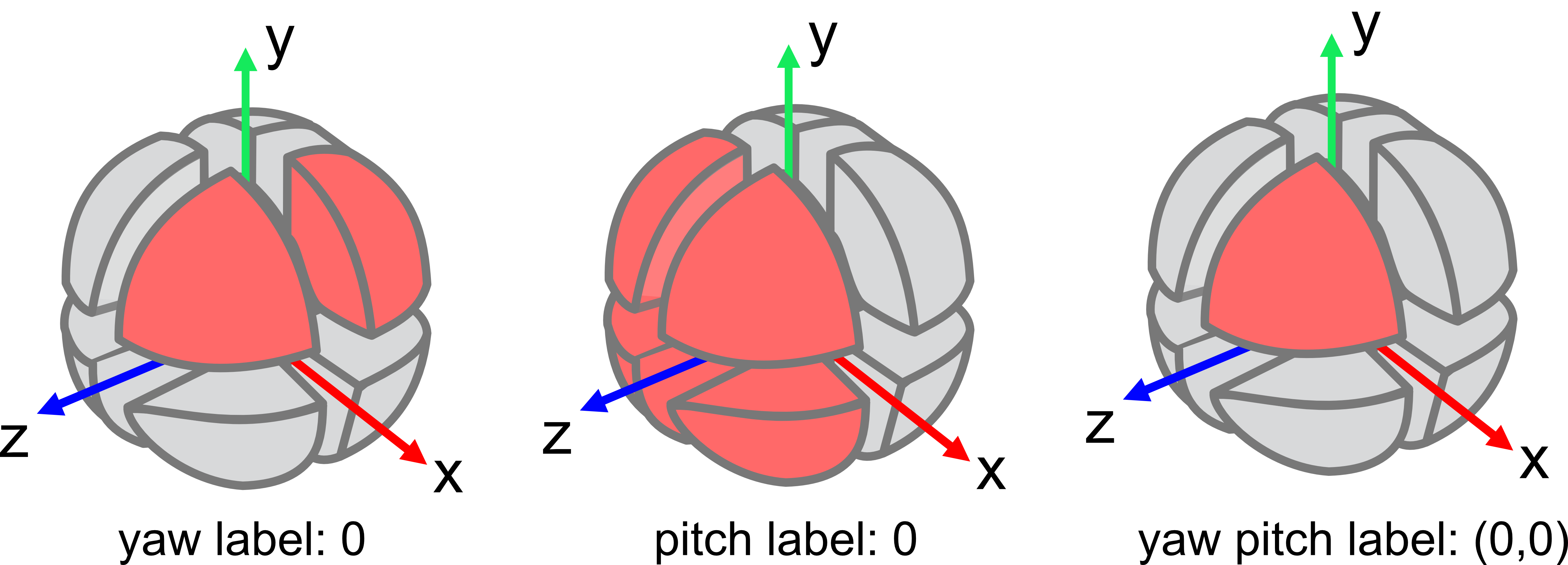}
    \captionof{figure}{An example of discretizing $\mathrm{SO}(3)$ into 8 bins ($\text{N}_\beta=2$, $\text{N}_\gamma=4$). Yaw and pitch labels are generated for $\gamma$ and $\beta$ in \figref{fig:yaw-pitch-cam}.}
    \vspace{-6pt}
    \label{fig:yaw-pitch-discretize}
\end{figure}

Unfortunately, designing \acp{dnn} that operates on sets is non-trivial. To alleviate this problem, we propose to discretize $\mathrm{SO}(3)$ into $\text{B}$ bins using the yaw and pitch of the Euler angles where each bin $\text{b}_{\text{i}} \in \text{B}$ is represented by a two-integer label $\text{b}_{\text{i}} = [\text{c}^{\text{i}}_{\beta},~\text{c}^{\text{i}}_{\gamma}] \in \mathbb{Z}_+^2$. The constraint $\text{C}$ then becomes one of the bins $\text{b}_\text{i} \in \text{B}$. Mathematically, the discretization is realized by first choosing a range of values for each of the two Euler angles pitch ($\beta \in [0,~\pi]$) and yaw ($\gamma \in [0,~2\pi]$). Next, these ranges are divided into equally spaced intervals of size $\text{N}_\beta$ and $\text{N}_\gamma$, respectively. Finally, a query Euler angle $\beta_\text{i}$ and $\gamma_\text{i}$ is mapped into the correct label $\text{b}_\text{i}=[\text{c}_\beta^{\text{i}},~\text{c}_\gamma^{\text{i}}]$ by:

\begin{equation}\label{eq:descritization}
\text{c}_{\beta}^{\text{i}} = \lfloor \frac{\text{N}_{\beta}\beta_{\text{i} }}{\pi} \rfloor,~\text{c}_{\gamma}^{\text{i}} = \lfloor \frac{\text{N}_{\gamma}\gamma_{\text{i}}}{2\pi} \rfloor.
\end{equation}
An example discretization using the above process is shown in \figref{fig:yaw-pitch-discretize}.

Using the two-integer constraint $\text{C}$, we design \methodname{} similarly to \cite{mousavian20196,lundell2023constrained} as a \ac{cvae}~\cite{sohnLearningStructuredOutput2015} but with both $\matr{O}$ \textit{and} \text{C} as the conditional variables. The \ac{cvae} consists of an encoder $\mathsf{q}_{\boldsymbol{\zeta}}(\mathbf{z}\mid\matr{O}, \text{C}, \matr{G})$ and a decoder $\mathsf{p}_{\boldsymbol{\chi}}(\matr{G}\mid\matr{O},\text{C},\mathbf{z})$, where $\mathbf{z}\in \mathbb{R}^{\text{L}}$ is a latent vector of size $\text{L}$. As we want to operate directly on \pcs{}, we choose \pointnet{} as the backbone of \methodname{}. The input \pc{} $\matr{X}\in \mathbb{R}^{\text{N}\times (3+\text{K})}$ to the encoder consist of the object \pc{} $\matr{O} \in \mathbb{R}^{\text{N}\times 3 }$ with the constraint $\text{C}$ and the grasp pose $\matr{G}$ as the K additional point-wise features. The decoder uses the same input \pc{} as the encoder but with $\mathbf{z}$ instead of $\matr{G}$ as additional point-wise features. 

For optimizing the parameters of \methodname{} we use the standard \ac{vae} loss:

\begin{align}
\label{eq:vae_loss}
    \mathcal{L}_{\text{VAE}} = \mathcal{L}(\matr{G}^*,\hat{\matr{G}}) + \eta \mathcal{D}_{\text{KL}}[\mathsf{q}_{\zeta} (\mathbf{z}\mid\matr{O},\text{C},\matr{G}^*),~\mathcal{N}(\matr{0},\matr{I})],
\end{align}
where $\eta$ is a scalar, $\matr{G}^*$ is a ground truth stable grasp, and $\mathcal{D}_{\text{KL}}$ is the KL-divergence. Similarly to \cite{mousavian20196,lundell2023constrained} we define the reconstruction loss $\mathcal{L}(\matr{G}^*,\hat{\matr{G}})$ as
\begin{align}
    \mathcal{L}(\matr{G}^*,\hat{\matr{G}}) = \norm{\text{h}(\matr{G}^*)-\text{h}(\mathbf{\hat{\matr{G}}})}_1,
\end{align}
where $\matr{\hat{G}}$ is a generated grasp from the decoder $\mathsf{p}_{\chi}$, and $\text{h}: \mathbb{R}^{7} \rightarrow  \mathbb{R}^{6\times 3}$ is a function that maps a 7-dimensional grasp pose into a \pc{} representation of the gripper $\matr{P} \in \mathbb{R}^{6\times 3}$. The benefit of a point cloud representation for the gripper pose is to combine orientation and translation into one loss function.

Although both the encoder and the decoder are used for training \methodname{}, only the decoder is used for sampling grasps on an unknown \pc{}. Specifically, to generate M grasps on an unknown object \pc{} $\matr{O}$ with the approach direction constrained to a yaw and pitch angle of $\beta_{\text{j}}$ and $\gamma_{\text{j}}$ the first step is to sample M iid latent vectors from the zero mean Gaussian $\mathbf{z}_{0,\dots,\text{M}}\sim \mathcal{N}(\matr{0},\matr{I})$.  Next, the yaw and pitch angles are mapped into the two-integer bin value $\text{b}_j$ using \eqref{eq:descritization}. Finally, M copies of the object \pc{} are created, where each \pc{} copy $\matr{O}_{\text{i}}$ is concatenated with a unique latent vector $\mathbf{z}_{\text{i}}$ and the constraint labels $\text{b}_{\text{j}}$, and passed through the decoder to produce the M grasps. 

\subsection{\ac{pc}-based Bin Selection}\label{sec:pc_selection}
As described above, the main downside of \methodname{} is the need to a-priori specify the approach direction constraint $\text{C}$. Two options for specifying $\text{C}$ are to use a human or to learn it. Unfortunately, both of these options are expensive to realize. Therefore, we propose selecting bins that align with some easy-to-estimate latent features of the object's geometry. 

The features we use in this work are the \acp{pc} of the object's point cloud as these have been used in similar non-learning-based grasping works discussed in \secref{sec:geometrically_constrained_grasping}. To calculate the \acp{pc}, we ultilize the \ac{pca} using the covariance matrix $\Sigma_{\matr{O}} = \frac{1}{N}(\matr{O}-\bar{\matr{O}})^{T}(\matr{O}-\bar{\matr{O}})$ of the object point cloud $\matr{O}$ with the barycenter $\bar{\matr{O}}$. This results in three ordered eigenvalues $\lambda_1$, $\lambda_2$, $\lambda_3$ and corresponding eigenvectors, $\hat{v}_1$, $\hat{v}_2$, $\hat{v}_3$, where $\lambda_1 \geq \lambda_2 \geq \lambda_3$. The eigenvectors are also known as the \acs{pc}. Following prior work \cite{balasubramanian2012physical}, we select the second largest \ac{pc} $\hat{v}_2$ to constrain the approach direction. 

Due to the symmetry of the \acp{pc}, we need to find the bins that intersect with $\pm \hat{v}_2$. For that, we first calculate the roll, pitch, and yaw separately for $\pm \hat{v}_2$ using

\begin{equation}\label{eq:vec_to_euler}
\begin{split}
&\alpha \in [0, 2\pi], \\
&\gamma = \arctan (\hat{v}_{2,y}, \hat{v}_{2,x}) \in [0, 2\pi], \\
&\beta = \arccos (\hat{v}_{2,z}) \in [0,\pi], \\
\end{split}
\end{equation}
where $\hat{v}_{2,x}$, $\hat{v}_{2,y}$, and $\hat{v}_{2,z}$ represent the x-, y-, and z-component of the vector $\hat{v}_{2}$ respectively.
Then, for each $\pm \hat{v}_2$, the pitch ($\beta$) and yaw ($\gamma$) angles are mapped into the corresponding two-integer label using \eqref{eq:descritization}. These two two-integer labels are then used one after the other to generate grasps with the approach vector constrained to $\pm \hat{v}_2$.    

\subsection{Grasp Discriminator}

\methodname{} can generate unsuccessful grasps between modes because it is only trained on distributions of successful grasps \cite{mousavian20196}. To avoid executing poor grasps, we used the grasp discriminator introduced in \cite{mousavian20196} to score how likely the sampled grasps were to succeed.

Mathematically, the  grasp discriminator $\mathcal{D}_{\boldsymbol{\psi}}(\text{S}=1 | \matr{G}, \matr{O})$ is optimized to approximates $\prob{(\text{S}=1 | \matr{G}, \matr{O})}$. It is based on the same \pointnet{} architecture as \methodname{}. However, the input \pc{} $\matr{Y} \in \mathbb{R}^{(\text{N}+6)\times (3+1)}$ to the evaluator differs significantly from the generator in that the object \pc{} $\matr{O} \in \mathbb{R}^{N \times 3}$ is concatenated with a grasp \pc{} $\matr{K} \in \mathbb{R}^{6\times 3}$ and an additional one-dimensional binary feature is added to the input \pc{} $\matr{Y}$ to distinguish between the two \pcs{} $\matr{O}$ and $\matr{K}$. The grasp discriminator is optimized to distinguish between successful and unsuccessful grasps using the binary cross-entropy loss:
\begin{align}
\mathcal{L}_{E}=(-\text{S}^*\log(\text{S})+(1-\text{S}^*)\log(1-\text{S})),    
\end{align}
where $\text{S}^*$ is the ground-truth success of a grasp and $\text{S}$ is the predicted success. 

\section{Dataset}\label{sec:dataset}

Training \methodname{} requires a dataset of successful grasps where the approach direction is labeled according to the discrete bins of $\mathrm{SO}(3)$. However, due to the discretization of $\mathrm{SO}(3)$ using the Euler angles, our dataset depends on which coordinate system we express the grasps in. Therefore, in this section, we first discuss the choice of the coordinate system and then introduce the dataset.    

\subsection{Camera-Centric Coordinate System}

Prior \ac{dl}-based grasping works have generally not discussed the coordinate system used for expressing their training data. The reason is that there is no need for such a discussion, as the training data can seamlessly be transformed from one coordinate system to another without influencing the grasp labels. Unfortunately, this is not the case in our work, as the labels C are coordinate-specific.

Two commonly used coordinate systems for expressing the grasp poses are the global and camera-centric coordinate systems. We choose to express the grasp poses in the camera-centric coordinate system because of two benefits. Firstly, in a camera-centric coordinate system, the object \pc{} is always aligned with the camera axes. Secondly, the camera-centric coordinate system enables a human-like way of communicating the grasp approach direction. For instance, with the camera-centric coordinate system, we can express grasp approach directions similar to a human from the left, right, above, below, behind, or in front of the object. Neither of these benefits applies when using a global coordinate frame unless the global coordinate system coincides with the camera coordinate system, which is rarely the case.

\subsection{The Approach-Constrained Grasping Dataset}

To train \methodname{}, we need a large-scale grasping dataset containing object \pcs{} $\matr{O}$ and successful grasps $\matr{G}$ where each grasp $\mathbf{g} \in \matr{G}$ has a specific constraint label C depending on its yaw and pitch angle. To date, no such grasping dataset exists. Still, instead of generating and labeling a completely new grasping dataset, we convert the grasps in the already established large-scale Acronym dataset~\cite{eppner2021acronym} that consists of 17.7 million simulated parallel-jaw grasps on 8872 objects from ShapeNet~\cite{chang2015shapenet} to fit our needs.

The steps for curating our dataset from the Acronym data are similar to the one presented in \cite{lundell2023constrained}, but with some key differences stemming from the need to discretize the space of rotations and label grasps accordingly. In detail, the steps we follow to create the dataset used in this work are:

\begin{enumerate}[label=(\roman*)]
    \item Discretize the space of rotations $\mathrm{SO}(3)$ into B bins using \eqref{eq:descritization}.
    \item For each object, randomize 100 camera poses pointing toward the object and render a \pc{} $\matr{O} \in \mathbb{R}^{\text{N}\times 3}$ from each pose as done in \cite{lundell2023constrained}.
    \item Transform 1000 random grasps $\matr{G}$ on each object from Acronym to the camera-centric coordinate system and then label them according to one of the B bins using \eqref{eq:vec_to_euler}.
\end{enumerate}
The above process was carried out on all the 8872 Acronym objects. From the final dataset, 112 random objects were held out for evaluation while the rest were used for training \methodname{}.



\section{experiment}\label{sec:exp}

We quantitatively evaluate the performance of \methodname{} in both simulated and real-world scenarios. Our evaluations aim to answer the following three key questions: 
\begin{enumerate}
\item How efficient is \methodname{} at sampling successful approach-constrained grasps compared to an unconstrained grasp sampler?
\item What is the impact on grasp success rate when sampling a direction bin based on the \ac{pc} direction over a random bin?
\item How successful is \methodname{} compared to an unconstrained generative grasp sampler in grasping real-world objects in an unconfined and confined space? 
\end{enumerate}

To answer these questions, we conducted three experiments, one in simulation and two in the real world. In the simulation, we benchmark \methodname{} with several discretization resolutions, while only one discretization resolution is used in the real-world experiments. \methodname{} is trained on \datasetacronomy{} with discretized labels, while \graspnet{} is trained on the same dataset without such labels. Both methods are trained with a latent space size $L=4$.

We use the grasp success metric, a binary variable evaluating 1 for a successful and 0 for an unsuccessful grasp, to quantify whether a grasp was successful. In all experiments, a grasp was deemed successful if the object was picked and remained within the gripper during a predefined movement. Moreover, to only evaluate grasps from specific directions, we removed all grasps outside the specified bins before evaluation. Consequently, if a particular direction is given and all generated grasps fall outside the specified bin, we deem that grasp trial unsuccessful.

\begin{figure}[tb]
    \centering
        \includegraphics[width=0.89\linewidth]{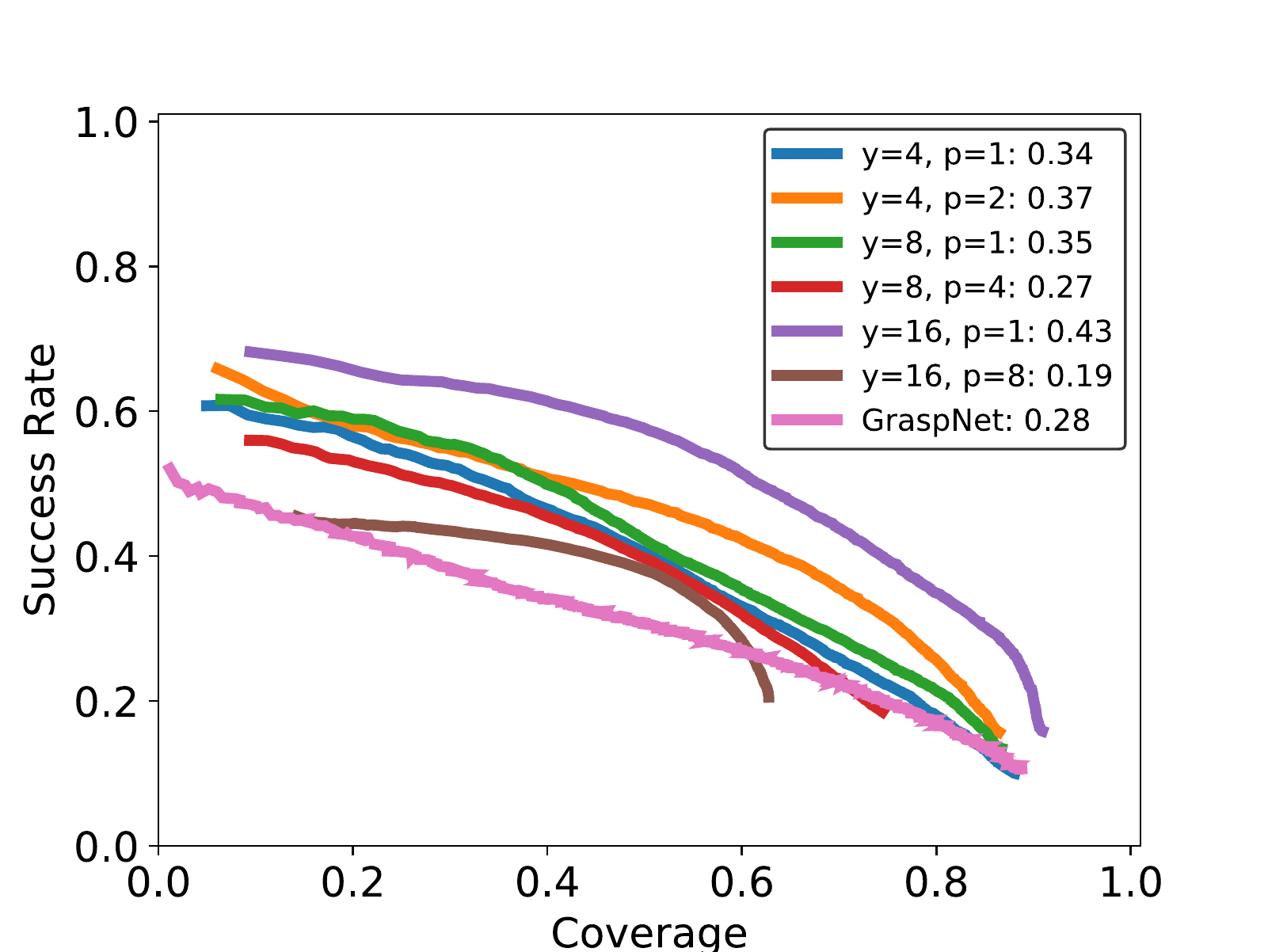}
    \caption{The comparison of varying discretization resolutions for yaw (y) and pitch (p). The \ac{auc} for each method is depicted in the image legend \bestcolor{}.}
    \label{fig:exp1_ablation_cov_suc}
\end{figure}


\subsection{Simulation Experiments}
In the simulation experiments, we used the Issac-Gym simulator \cite{makoviychuk2021isaac} to evaluate all methods. In this simulator, we initialized the test object as free-floating, used a side-facing depth camera to capture \pcs{}, and a virtual 7-DoF Franka Panda robot for grasping. Gravity was activated after the object was grasped. 

Besides evaluating the grasp success rate, we also assessed the success-over-coverage rate in simulation, a metric originally proposed for grasping in \cite{mousavian20196}. To assess the coverage, we first discretized $\mathrm{SO}(3)$ for each test object into 128 so-called ground-truth bins. Then, we associated each ground-truth grasp for each object with the correct ground-truth bin based on the approach direction, forming 128 different grasp subsets $\matr{G}^{*}=\{\matr{G}_{1}^*,~\matr{G}_{2}^*,~\dots,~\matr{G}_{128}^*\}$. Bins with no ground-truth grasps were ignored. These split allowed us to evaluate grasp coverage between grasps from ground-truth bins that overlap with the bin in \methodname{} that grasps were generated from. A generated grasp covers a ground-truth one if the angle between their approach vectors is less than 10 degrees and the translation distance is less than 2 cm. 

We first compared different discretization resolutions in the simulator to study their effect on the success-over-coverage rate. To this end, we evaluated the following 6 \methodname{} models with varying numbers of yaw and pitch (y/p) bins: (4/1), (4/2), (8/1), (8/4), (16/1), and (16/8). For each object, 1000 grasps were sampled and scored by the discriminator.

The results are presented in \figref{fig:exp1_ablation_cov_suc}. These results clearly indicate that \methodname{} with all resolutions achieve a much higher success-over-coverage rate than the \graspnet{} baseline. The reason \graspnet{} scores a much lower success-over-coverage rate is that it generates grasps all over the object, while \methodname{} leveraged orientation information to only generate approach-constrained grasps in the bins. We observed that \methodname{} with large pitch resolutions generally performed worse than the other methods. One reason for this is that the pitch bins could be placed in occluded regions on the object \pc{}, and grasps sampled from such directions with large pitch resolutions naturally had lower quality and tended to align with the mean grasping pose from the training data. Interestingly, \methodname{} (16/8) with the highest pitch and yaw resolution performed even worse than \graspnet{}, which we hypothesize stems from many low-quality grasps sampled from the occluded regions. 

\begin{figure}
	\centering
        \includegraphics[width=1\linewidth]{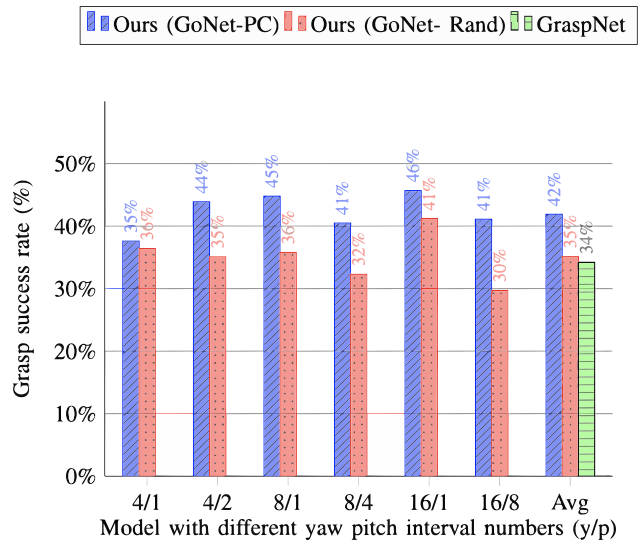}
	\vspace{-1em}
	\caption{The per model grasp success rates with varying yaw and pitch (y/p) resolution. The last column shows the average grasp success rate for all constrained models and \graspnet{}.}
	\label{fig:simresult_suc_rate}
\end{figure}

Finally, we compared the success rates of \methodname{} with the \ac{pc}-based bin selection, random bin selection, and \graspnet{}. We sampled and evaluated 400 grasps for each model and executed the 20 best grasps according to their grasp score. The results, presented in \figref{fig:simresult_suc_rate}, demonstrate that \methodname{} with the \ac{pc}-based bin selection consistently outperforms the other models, achieving, on average, a 7\% higher grasp success rate than a random bin and an 8\% better success rate than \graspnet{}, highlighting, once again, the effectiveness of generating grasps from geometrically meaningful approach directions as discussed in \cite{balasubramanian2012physical}. Interestingly, the average success rate of \methodname{} with random bin selection is on par with \graspnet{}, indicating that our method can find good grasps even from random approach directions. 

\subsection{Real-world Experiments}

\begin{figure}[tb]
    \centering
        \includegraphics[width=0.9\linewidth]{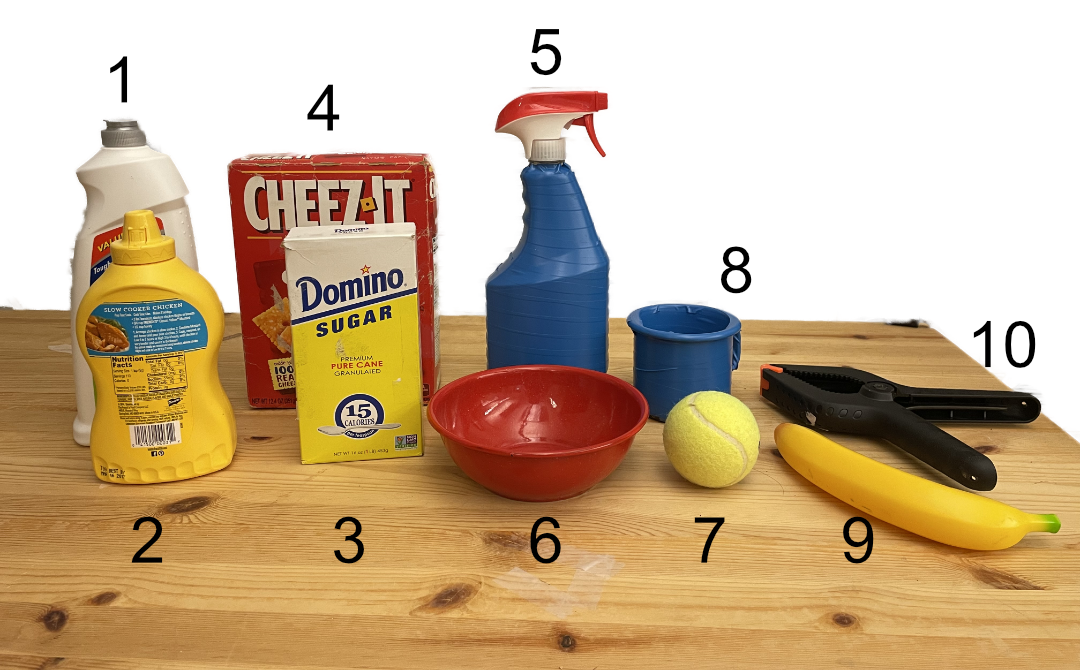}
    \caption{Experimental objects. From 1 to 10: bleach cleanser, mustard bottle, sugar box, cracker box, Windex bottle, metal bowl, tennis ball, metal mug, banana, and spring clamp.}
    \label{fig:full_object_set}
\end{figure}

\begin{figure}[tb]
    \centering
        \includegraphics[width=1\linewidth]{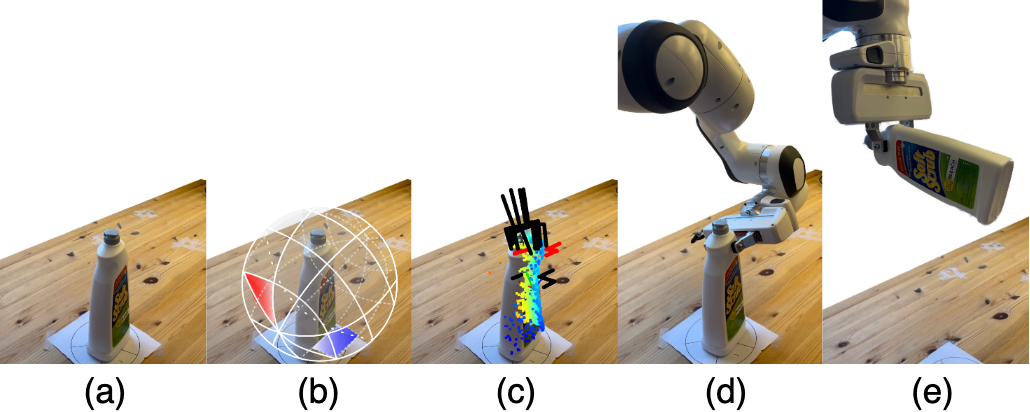}
    \caption{An example of \methodname{} generating and picking the bleach cleanser from the table. (a) The object is placed in a specific pose. (b) The $\mathrm{SO}(3)$ space around the object is discretized into a pre-defined yaw and pitch resolution where the red bin is the \ac{pc} one, and the blue is the top-down one. (c) The sampled grasps with the highest-scoring one in red. (d) The robot successfully reaches the red grasp pose, and (e) moves back to its start pose.}
    \label{fig:table_exp_pca}
\end{figure}


\begin{figure}[tb]
    \centering
        \includegraphics[width=1\linewidth]{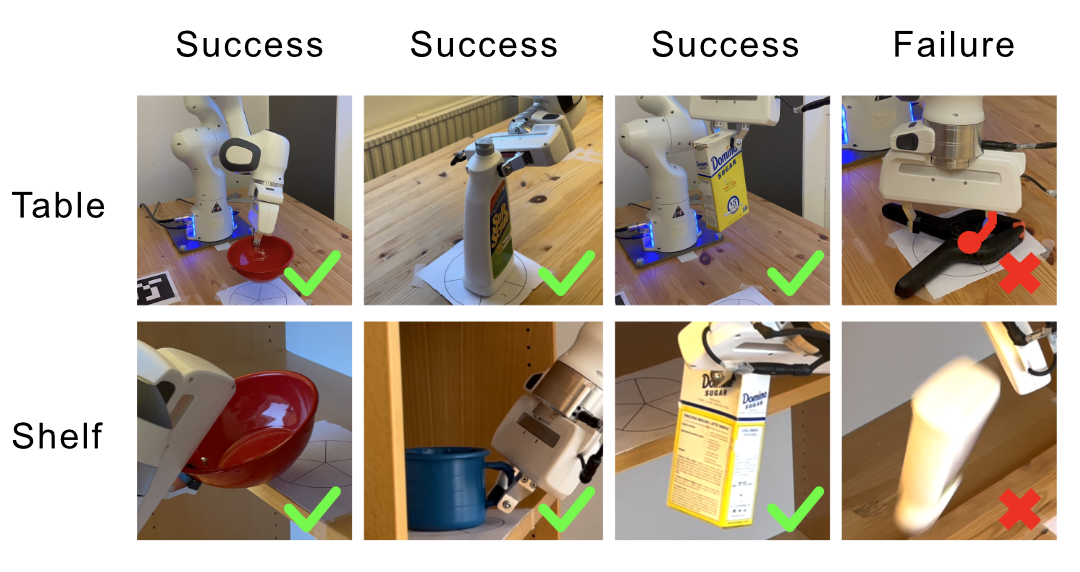}
    \caption{Some successful and failed grasp examples for table-picking (top row) and shelf-picking (bottom row). The first three columns depict successful grasps, while the last column shows two different failure cases: collision (top image) and slip (bottom image).}
    \label{fig:suc-fail-case}
\end{figure}

\begin{table}[ht]
    \centering
    \begin{adjustbox}{max width=\linewidth}
    \setlength{\tabcolsep}{4pt}
         \begin{tabular}{lccccccc}
            \toprule
            \textbf{Experiment}& \multicolumn{4}{c}{\textbf{Table-picking}} & \multicolumn{3}{c}{\textbf{Shelf-picking}}\\
            \cmidrule(l){2-5} \cmidrule(l){6-8}
            Object & GN & GN-tp & \methodname{}-t & \methodname{}-tp & GN & GN-cf & \methodname{}-cf\\ \midrule
            1. Bleach cleanser & 4/5 & 1/5  & 1/5   & 4/5  & 0/5  & 3/5 & 3/5 \\ 
            2. Mustard bottle  & 1/5 & 1/5  & 1/5   & 3/5  & 1/5  & 2/5 & 1/5 \\ 
            3. Sugar box       & 5/5 & 4/5  & 5/5   & 5/5  & 0/5  & 1/5 & 4/5 \\ 
            4. Cracker box     & 4/5 & 5/5  & 5/5   & 5/5  & 1/5  & 0/5 & 3/5 \\
            5. Windex bottle   & 4/5 & 4/5  & 5/5   & 5/5  & 0/5  & 1/5 & 0/5 \\
            6. Metal bowl      & 5/5 & 5/5  & 5/5   & 5/5  & 1/5  & 5/5 & 4/5 \\
            7. Tennis ball     & 1/5 & 4/5  & 4/5   & 5/5  & 1/5  & 0/5 & 3/5 \\
            8. Metal mug       & 5/5 & 5/5  & 5/5   & 5/5  & 3/5  & 3/5 & 3/5 \\
            9. Banana          & 0/5 & 2/5  & 3/5   & 3/5  & 0/5  & 0/5 & 0/5 \\
            10. Spring clamp    & 0/5 & 2/5  & 2/5   & 2/5  & 0/5  & 0/5 & 0/5 \\
            \midrule 
            Avg. success rate  &58\%&66\% &72\% & \textbf{84\%}  & 14\% & 30\% & \textbf{42\%}\\
            Ratio grasps kept    & -- & 8.63\% &99.80\% &100\%  & -- & 13.28\% & 99.59\%\\
            \bottomrule
        \end{tabular}
    \end{adjustbox}
    \caption{Results for table-picking and shelf-picking. GN-tp denotes \graspnet{} with top-down and \ac{pc}-based filtering, \methodname{}-t denotes \methodname{} with top-down only grasping, \methodname{}-tp denotes \methodname{} with top-down and \ac{pc}-based grasping, GN-cf denotes \graspnet{} with collision-free filtering and \methodname{}-cf denotes \methodname{} with collision-free bin grasping.}
    \label{result_exp}
    \end{table}

In the real-world experiments, we conducted two experiments: a table-picking experiment and a shelf-picking experiment. The table picking experiments shown in \figref{fig:table_exp_pca} is an example of an unconfined picking environment, while the shelf picking experiment is an example of a confined picking environment. In both experiments, we used a 7-\ac{dof} Franka Panda robot for grasping and a Kinect V2 for capturing \pcs{} of the objects. An Aruco marker \cite{arucomarker} placed on the table was used for extrinsic camera calibration.
The objects used in the experiments are shown in \figref{fig:full_object_set} and were selected due to their diverse geometries. All these objects except the blue metal mug were from the YCB dataset~\cite{calli2017yale}. In both experiments, each object was placed on the plane in five equally distributed orientations (0\textdegree, 72\textdegree, 144\textdegree, 216\textdegree, and 288\textdegree). For each object orientation, we sampled 200 grasps and removed all that would collide with the environment. We then scored the remaining grasps using the grasp discriminator and executed the first reachable grasp of the top 10 scoring grasps. In total, this amounts to 50 grasp trials per method per experiment. A grasp trial was successful if the object remained within the gripper after the robot returned to the starting position and rotated the hand by ±90 degrees. If the object dropped or no reachable grasps were found, the trial failed.

We compare the 8/4 \methodname{} from the simulation experiments against two \graspnet{} versions: one where all sampled grasps were kept, and another one where all grasps with approach directions outside the bins specified by \methodname{} were removed. In both real-world experiments, we evaluated the grasp success rate and the ratio of grasps kept, which measures the sample efficiency of the methods. In the table-picking experiment, the bins aligned with the second \ac{pc} of the object \pc{} and the top-down bin were selected. In the shelf-picking experiment, we selected the bins in \figref{fig:shelf_exp_with_grasps} that would result in collision-free grasps.

The results for both experiments are presented in Table \ref{result_exp}, and some example grasps are shown in \figref{fig:suc-fail-case}. Overall, \methodname{} outperformed \graspnet{}, with an 18\% higher average grasp success rate on table-picking and 12\% higher on shelf-picking. We attribute this improvement to \methodname{}'s ability to efficiently generate successful grasps in specific directions, highlighted by a 99.5\% ratio of grasps kept compared to less than 13.3\% for \graspnet{}. 

It is worth highlighting that the grasp success rates of our experimental results are significantly lower than those reported in \cite{mousavian20196}. This difference can be attributed to three factors. First, including rotational and linear acceleration tests increases the risk of object slippage, as depicted in Figure \ref{fig:suc-fail-case}. Second, we validate the sampling methods without incorporating grasp refinement, which likely reduces success rates further. Lastly, our side-viewing camera also impacts the success rate negatively, which aligns with the findings in \cite{lundell2020beyond} where the effect viewing angle has on grasp success was investigated.

Another interesting finding is that \methodname{} with top-down and \ac{pc}-based bins achieved a higher success rate on the Bleach cleanser and Mustard bottle in table picking than when the top-down only bin was used. For such objects, the grasps approaching along the second \ac{pc} are more stable than the top-down direction. However, the combination of \ac{pc} and top-down does not improve the success rate on box-shaped objects, as both approach directions are good for grasping such objects. 

\section{conclusion}

We proposed \methodname{}, the first deep learning-based data-driven grasp sampler that can generate approach-constrained grasps in all of $\mathrm{SO}(3)$ together with a geometrical method to automatically extract specific grasp approach directions from the object's \pc{}. In simulation and real-world experiments, \methodname{} reached an 8--18\% higher grasp success rate and kept more than 7 times as many grasps as an unconstrained data-driven grasp sampler. The real-world shelf-picking experiment highlighted \methodname{}'s ability to pick from confined spaces by constraining grasp approach directions to collision-free directions instead of sampling all around the object.

In the future, we envision \methodname{} to constrain complete grasp poses to subsets of $\mathrm{SE}(3)$, including subsets that are easy to reach and have high manipulability. However, for that to become a reality, we believe that the discrete constraint needs to be replaced with a continuous one and that the orientation constraint should be combined with the position constraint from \cite{lundell2023constrained}, both of which are non-trivial but exciting future work directions.  



\bibliographystyle{IEEEtran}
\bibliography{ref}

\end{document}